\documentclass[conference]{IEEEtran}
\usepackage{cite}

\ifCLASSINFOpdf
  \usepackage[pdftex]{graphicx}
\else
\fi
\ifCLASSOPTIONcompsoc
 \usepackage[caption=false,font=normalsize,labelfont=sf,textfont=sf]{subfig}
\else
 \usepackage[caption=false,font=footnotesize]{subfig}
\fi
\usepackage{url}

\usepackage{amsmath,amssymb} 
\usepackage{booktabs}		
\usepackage{xspace}		
\usepackage[]{units}		
\usepackage{tabularx}
\usepackage{array}
\usepackage{siunitx}		
\makeatletter
\DeclareRobustCommand\onedot{\futurelet\@let@token\@onedot}
\def\@onedot{\ifx\@let@token.\else.\null\fi\xspace}

\def\eg{\emph{e.g}\onedot} 
\def\ie{\emph{i.e}\onedot}

\makeatother

\usepackage[final]{changes}

\usepackage{todonotes}

\usepackage{hyperref}	

\begin{document}
%
\title{PetroSurf3D -- A Dataset for high-resolution 3D Surface Segmentation}



%
\author{\IEEEauthorblockN{Georg Poier\IEEEauthorrefmark{1},
Markus Seidl\IEEEauthorrefmark{2},
Matthias Zeppelzauer\IEEEauthorrefmark{2}, 
Christian Reinbacher\IEEEauthorrefmark{1}, \\
Martin Schaich\IEEEauthorrefmark{3}, 
Giovanna Bellandi\IEEEauthorrefmark{4}, 
Alberto Marretta\IEEEauthorrefmark{5} and
Horst Bischof\IEEEauthorrefmark{1}}
\IEEEauthorblockA{\IEEEauthorrefmark{1}Graz University of Technology}
\IEEEauthorblockA{\IEEEauthorrefmark{2}St. P\"olten University of Applied Sciences}
\IEEEauthorblockA{\IEEEauthorrefmark{3}ArcTron 3D}
\IEEEauthorblockA{\IEEEauthorrefmark{4}University of Cambridge}
\IEEEauthorblockA{\IEEEauthorrefmark{5}Parco Archeologico Comunale di Seradina-Bedolina}}


\maketitle

\begin{abstract}


The development of powerful 3D scanning hardware and reconstruction algorithms has strongly promoted the generation of 3D surface reconstructions in different domains. An area of special interest for such 3D reconstructions is the cultural heritage domain, where surface reconstructions are generated to digitally preserve historical artifacts. While reconstruction quality nowadays is sufficient in many cases, the robust analysis (e.g. segmentation, matching, and classification) of reconstructed 3D data is still an open topic. In this paper, we target the automatic and interactive segmentation of high-resolution 3D surface reconstructions from the archaeological domain. To foster research in this field, we introduce a fully annotated and publicly available  large-scale 3D surface dataset including high-resolution meshes, depth maps and point clouds as a novel benchmark dataset to the community. We provide baseline results for our existing random forest-based approach and for the first time investigate segmentation with convolutional neural networks (CNNs) on the data. Results show that both approaches have complementary strengths and weaknesses and that the provided dataset represents a challenge for future research.

\end{abstract}


%
\IEEEpeerreviewmaketitle

\section{Introduction}

Today, powerful techniques \deleted{exist }for the reconstruction of 3D surfaces 
\replaced{exist, such as}{ (e.g.} laser scanning, structure from motion\replaced{ and}{,} 
structured light scanning\deleted{)} \cite{wu2013}. 
The result is an increased availability of surface reconstructions with high resolutions 
\replaced{at}{going down to millimeter and even} sub-millimeter scale. 
At these high resolutions it is possible to capture the geometric fine structure 
(i.e. the topography \cite{blunt2003}) of a surface. 
The surface topography determines the \textit{tactile appearance} of a surface and is thus 
characteristic for different materials and differently rough surfaces. 
The automatic segmentation and classification of surfaces 
according to their topography 
\added{is an essential pre-requisite for reliable large scale analyses, 
however, it} is still an open problem.

A crucial requirement for the development of automatic surface segmentation
algorithms are publicly available datasets with
precise manual annotations (ground truth). 
%
%
A large number of datasets has been published for 2D and 3D 
texture analysis and material classification 
\cite{Dana1999tog,Ojala2002icpr,Haindl2008icpr}.
Usually, no geometric information is provided with these datasets
\ie, the datasets contain only images of the surfaces
(potentially with different lighting directions).
Automatic segmentation methods, however, are supposed to benefit strongly
from full 3D geometric information compared to only 2D (RGB) texture. Other datasets, employed for semantic segmentation, indeed provide 3D information
\cite{firman2016cvprw,Armeni2016arXiv,Song2015cvpr,Silberman2012eccv,Janoch2011iccvw,Silberman2011iccvw} but at a completely different spatial scale.
These datasets are usually captured
using off-the-shelf depth cameras (e.g. Microsoft Kinect) and have primarily been developed for
scene understanding and object recognition. Thus, they show entire 
objects and scenes and provide resolutions at centimeter level. 
These datasets address a different task and are too coarse to capture the characteristics of different types of surfaces and materials. 

In this paper, we present a dataset of high-resolution 3D surface reconstructions which contains full geometry information as well as color information and thus resembles both the tactile and visual appearance of the surfaces at a micro scale. The surfaces stem from the archaeological domain and represent natural rock surfaces \replaced{into which}{where} petroglyphs (i.e. symbols, figures and abstractions of objects) have been pecked, scratched or carved \replaced{in}{into during} ancient times. The engraved motifs represent areas with different roughness and tactile structure and exhibit complex and heterogeneous shapes. \replaced{H}{Due to h}undreds or \added{in most cases even} thousands of years of weathering and erosion \replaced{rendered many } {the }petroglyphs \deleted{are often }indistinguishable
from the natural rock surface \replaced{with the naked eye or by using}{ from pure} 2D imagery. These properties make the scanned surfaces a challenging testbed for the evaluation of automatic 2D and 3D surface segmentation algorithms.



This paper builds upon a series of incremental previous works on 
3D surface segmentation and 
classification \cite{mzz15, Zeppelzauer2015dh, zeppelzauer2016jooch} and 
intends to consolidate and extend the achieved results. 
Our contribution beyond previous research are as follows:
\begin{itemize}
	\item We present a novel benchmark dataset for surface segmentation of high-resolution 3D surfaces to the public that enables objective comparison between novel surface segmentation techniques.
	\item We provide precise ground truth annotations generated by experts from archeology for the evaluation of surface segmentation algorithms together with a reproducible evaluation protocol.
	\item We provide baselines for our existing approach \cite{Zeppelzauer2015dh, zeppelzauer2016jooch} and a novel CNN-based approach to enable instant performance comparisons. 
	\item \replaced{We comprehensively evaluate}{A comprehensive evaluation of} the generalization ability of the proposed approaches and the benefit of using full 3D information for segmentation compared to pure 2D texture segmentation.
\end{itemize}

\section{Dataset}
\label{sec:dataset}

\replaced{Our}{To fill this gap, we created a} 
fully labeled 3D dataset of rock surfaces with petroglyphs
is publicly available\footnote{\url{
http://lrs.icg.tugraz.at/research/petroglyphsegmentation/}}. 
In a large effort, we scanned petroglyphs on several different
rocks at sub-millimeter accuracy.
From the 3D scans we created meshes and point clouds and additionally 
orthophotos and corresponding depth maps
to enable the application of 3D and 2D segmentation approaches 
on the data. Note that, since there are usually no self-occlusions
in pecked rock surfaces, the 3D information is \added{almost fully} preserved in the
depth maps (except for rasterization artifacts).
For all depth maps and orthophotos we provide pixel-wise ground truth labels
(overall about 232 million labeled pixels) and the parameters for the mapping from 3D space to 2D (and vice versa).

\subsection{Dataset Acquisition}

The surface data has been acquired \deleted{in Summer 2013 }at the UNESCO World 
heritage site in Valcamonica, Italy, which provides one of the largest 
collections of rock art in the world\footnote{http://whc.unesco.org/en/list/94, 
last visited February 2017}. 
\deleted{The surfaces to be scanned have been carefully selected  
by archaeologists with the intention to 
maximize diversity across different styles, 
shapes, scenes, and locations. 
Furthermore, regions which were never scanned in 
3D before were given preference.} The data has been scanned 
by experts 
using two different scanning techniques: (i) structured light scanning (SLS) 
with the Polymetric PTM1280 scanner in combination with the associated software 
QTSculptor and (ii) structure from motion (SfM). For SfM, photos were acquired 
with a high-quality Nikkor \unit[60]{mm} macro lense mounted on a Nikon D800. 
For bundle adjustment the SfM engine of 
the software package Aspect3D\footnote{http://aspect.arctron.de, last visited February 2017} 
was used and 
SURE\footnote{http://www.ifp.uni-stuttgart.de/publications/software/sure/index.en.html, 
last visited February 2017} was employed for the densification of the point clouds. 
The point clouds have been 
denoised by removing outliers which stand out significantly from the 
surface~\cite{rusu2008towards}
and smoothed by a moving least squares filter\footnote{Both filters 
are implemented in the Point Cloud Library (PCL) http://pointclouds.org, 
last visited February 2017}. 
The resulting point clouds have a sampling distance of at least \unit[0.1]{mm} and provide 
RGB color information for each 3D vertex. The vertex coordinates are in metric units 
relative to a base station. We provide the point clouds 
in XYZRGB format\replaced{.}{, which is an ASCII format where every line contains one point 
of the cloud: $<X, Y, Z, R, G, B>$.}
Additionally, the point clouds were meshed by Poisson triangulation.
Meshes were textured with the captured vertex colors and are provided in WRL format.



\replaced{We generated}{For the derivation of} orthophotos and depth maps \replaced{of all surface reconstructions.}{ we estimate a support plane for the 
input mesh by estimating a median plane from a subset of its 
points. Next, we estimate the location of each 3D point on the 
support plane by projecting the point along the normal direction of the plane.
We map the signed distances between the 3D points and the 
plane to the respective projected 
location on the plane. The result is a 2D depth map of the 3D 
surface. Similarly, the orthophoto is generated by mapping the
RGB colors to the support plane.} For the rasterization of the 
projected images \added{we used} a resolution of 300dpi (\ie, \unit[0.08]{mm} pixel side length). 
\replaced{The ortho projections were derived from the meshed 3D data}{We used meshes to generate the projections} since \replaced{this enables}{, this way,} 
a dense projection without holes\deleted{ is possible}.
The depth maps are stored as 32-bit TIFF files.

For each surface a pixel-accurate ground truth has been generated 
by archaeologists who labeled all pecked regions on the surface. 
Since the surfaces contain no self-occlusions the annotators 
worked directly on the 2D orthophotos and depth maps. 
\deleted{For this purpose an image processing program with different brush tools 
was used to produce the ground truth annotation.}
The annotators spent several hours on each surface 
depending on the size and complexity of the depicted engraving, e.g. anthropomorph, inscription, symbol, etc.
Anthropogenically altered, \ie pecked, areas were annotated with white color,
whereas the natural rock surface 
remained black and regions outside the scan were colored red.
\deleted{The archaeologists reported that -- besides the tedious procedure -- 
they sometimes experienced difficulties in 
annotating pecked regions from the orthophotos due to their similar visual 
appearance to the natural rock surface.}
The provided geometric mapping information between the 3D point cloud and the ortho projections allows to easily map the ground truth to the point cloud and the mesh for processing in the 3D space.

\subsection{Dataset Overview}

The final dataset contains 26 high-resolution surface reconstructions of 
natural rock surfaces with a large number of petroglyphs.
Tab.~\ref{tab:DataStatistics} provides some basic measures for each 
reconstruction, such as number of points, covered area, percentage of 
pecked surface area etc.
The petroglyphs have 
been captured at various locations at three different sites 
in the valley: ``Foppe di Nadro" (IDs 1-3), ``Naquane" (IDs 4-10), and 
``Seradina" (IDs 11-26).
\deleted{We list the surface reconstructions contained in the dataset in 
Tab.~\ref{tab:MaterialOverview}. Each of the three sites is 
partitioned into different \emph{rocks} and larger rocks are further 
subdivided into multiple \emph{areas}. }
The point clouds of all surfaces together 
sum up to overall 115 million points. They cover in total an area 
of around \unit[1.6]{m\textsuperscript{2}}. 
After projection to orthophotos 
and depth images this area corresponds to 
around 232 million pixels.
Note that there are more pixels than 3D points 
due to the interpolation that takes place during projection of the mesh.

\begin{table}[t]
	\tiny
	\centering
	\caption{Overview of basic measures of the digitized surfaces: 
	the covered area (in pixels at 300dpi and in cm\textsuperscript{2}), 
	the number of 3D points in the point cloud, 
	the percentage of pecked regions, 
	the number of disconnected pecked regions, 
	the range of depth values}
                {
     \begin{tabular}{c r r r S[table-format=3.2] S[table-format=3.0] S[table-format=3.2]}
		\toprule
			\textbf{ID}	& \multicolumn{2}{c}{\textbf{$\ \ \ \ \ $Covered Area}} & \textbf{Num.}   & \textbf{Percentage} & \textbf{Num.} & \textbf{Depth Range} \\
			& \textbf{in px} & \textbf{in cm\textsuperscript{2}} &	\textbf{3D Pts.} &	\textbf{Pecked} &	\textbf{Seg.} & \textbf{in mm} \\
		\midrule
                        1  & 5 143 296  &  368.69 &   3 264 005 & 14.61 & 48 & 2.89     \\
                        2  & 15 638 394  &  1121.03 & 10 280 976 & 10.56 & 21 & 4.83   \\
                        3  & 8 846 214  &  634.14 &   5 503 742 & 47.63 & 18 & 9.11     \\
                \midrule
                        4  & 15 507 622  &  1 111.66 & 3 782 381 & 14.96 & 17 & 62.52  \\
                        5  & 16 994 561  &  1 218.25 & 2 658 330 & 17.27 & 44 & 70.60  \\
                        6  & 13 102 254  &  939.23 &  1 260 401 & 12.67 & 13 & 49.32   \\
                        7  & 12 035 386  &  862.75 &  810 312  & 34.02 & 26 & 15.17   \\
                        8  & 12 834 446  &  920.03 &  8 677 163 & 26.17 & 45 & 6.74    \\
                        9  & 12 835 586  &  920.11 &  8 386 259 & 32.83 & 29 & 3.82    \\
                        10 & 5 901 454  &  423.04 &   2 096 476 & 21.59 & 9 & 5.41      \\
                \midrule
                        11 & 5 632 144  &  403.74 &   3 541 799 & 9.26 & 23 & 10.23     \\
                        12 & 7 103 936  &  509.24 &   4 432 013 & 5.09 & 6 & 10.22      \\
                        13 & 6 155 628  &  441.26 &   3 810 000 & 8.26 & 63 & 19.85     \\
                        14 & 5 855 280  &  419.73 &   4 417 779 & 6.47 & 17 & 10.50     \\
                        15 & 4 855 764  &  348.08 &   2 981 570 & 4.44 & 24 & 9.39      \\
                        16 & 4 029 231  &  288.83 &   2 523 543 & 6.58 & 29 & 4.27      \\
                        17 & 4 838 487  &  346.84 &   3 022 433 & 3.15 & 27 & 21.75     \\
                        18 & 6 396 152  &  458.50 &   4 007 232 & 19.41 & 25 & 9.45     \\
                        19 & 7 141 253  &  511.92 &   4 472 845 & 18.20 & 32 & 17.32    \\
                        20 & 6 864 476  &  492.08 &   4 238 990 & 12.02 & 15 & 21.39    \\
                        21 & 3 909 579  &  280.26 &   2 255 030 & 20.40 & 61 & 5.32     \\
                        22 & 4 073 804  &  292.03 &   2 395 125 & 16.34 & 65 & 3.99     \\
                        23 & 3 612 131  &  258.93 &   2 113 670 & 24.23 & 54 & 5.33     \\
                        24 & 19 104 798  &  1 369.52 & 10 685 564 & 26.61 & 152 & 27.35 \\
                        25 & 14 920 005  &  1 069.53 & 8 188 025 & 15.55 & 63 & 17.49  \\
                        26 & 8 921 684  &  639.55 &   5 515 973 & 15.59 & 99 & 16.62    \\
		\midrule
			Overall & 232 253 565  &  16 648.97 & 115 321 636 & 18.68 & 1 025 & \text{[2.89, 70.60]}\\
		\bottomrule
     \end{tabular}%

	}
	\label{tab:DataStatistics}
\end{table}

The scans show isolated figures 
as well as scenes with multiple interacting petroglyphs (e.g. hunting scenes).
The pecked regions in all reconstructions are disconnected and in average 
consist of about 40 segments. The pecked regions make up 
around 19\% of the entire scanned area. 

\begin{figure*}[t]
\centering
 \subfloat
  {\includegraphics[width=0.31\textwidth]{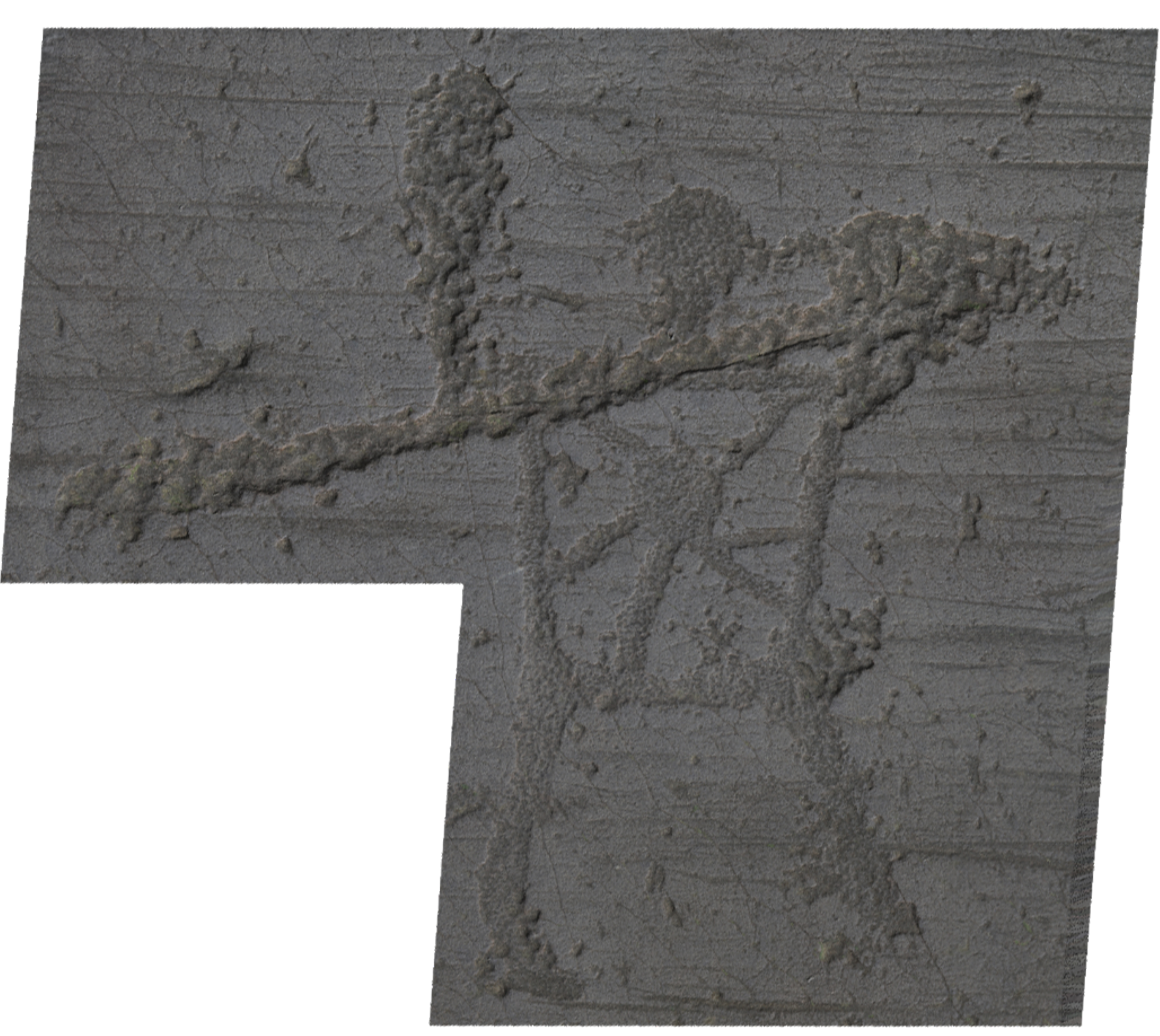}}
 \quad
 \subfloat
  {\includegraphics[width=0.31\textwidth]{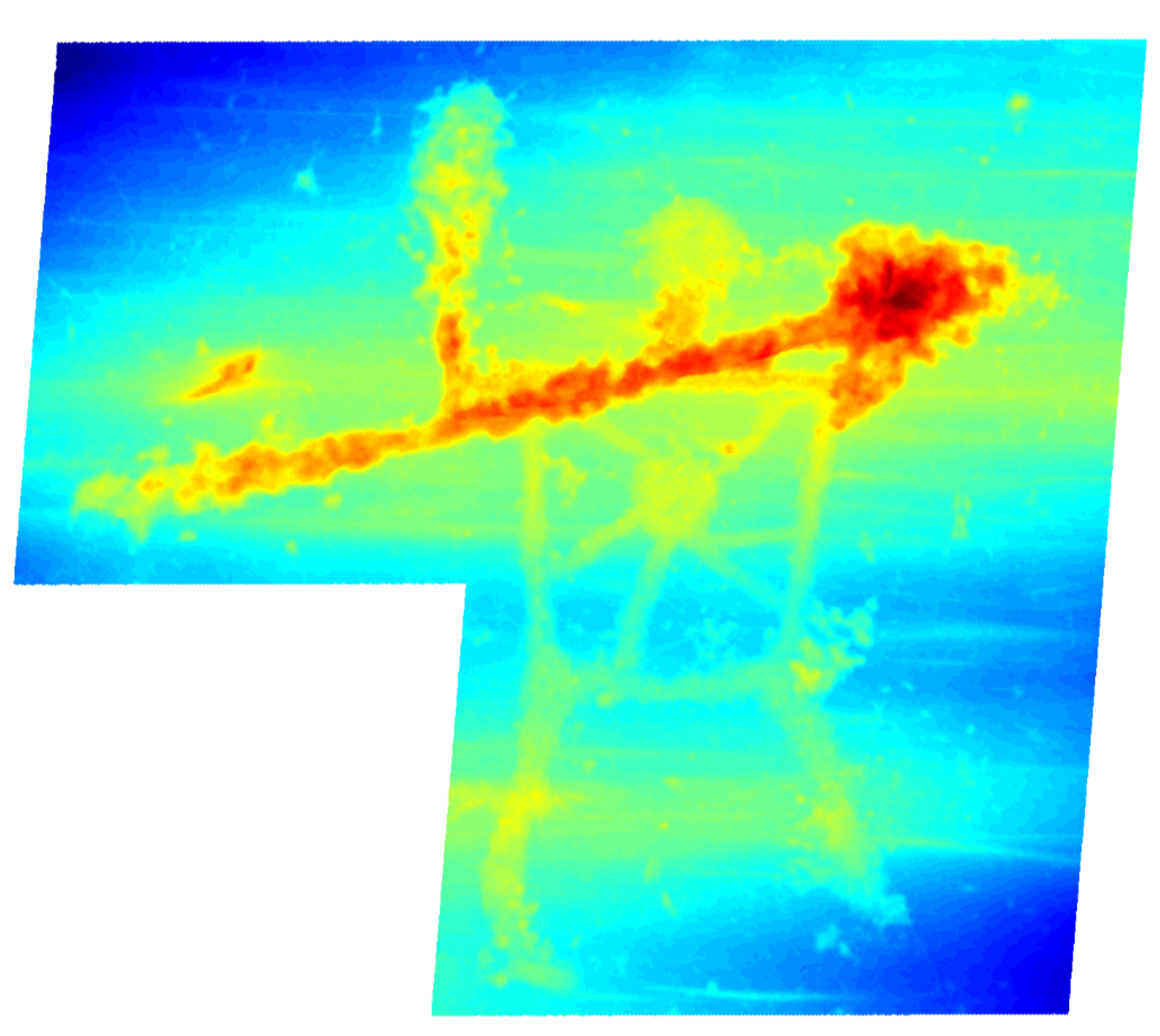}}
 \quad
 \subfloat
  {\includegraphics[width=0.31\textwidth]{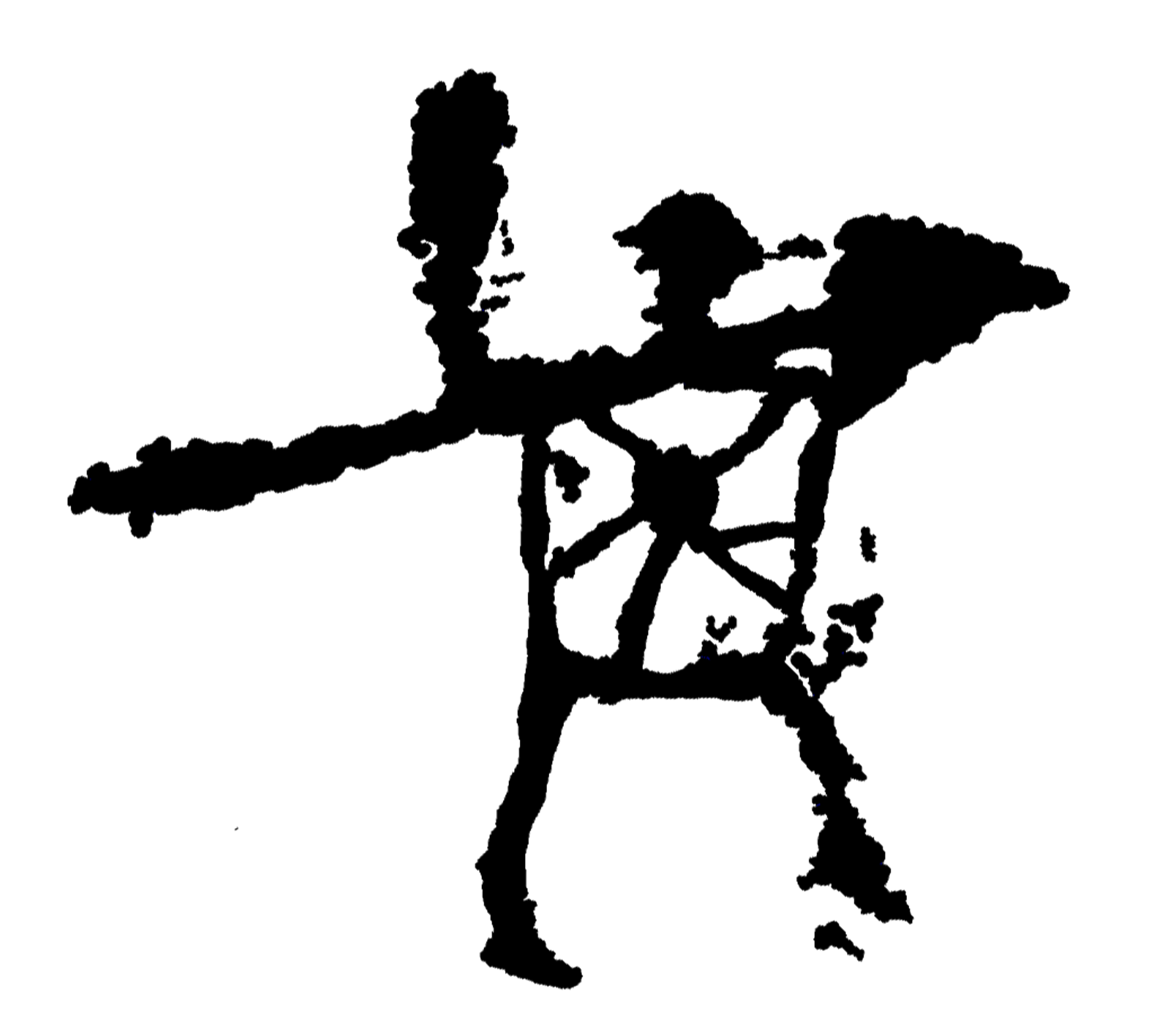}}
 \\
 \caption{
 Example orthophoto (left), corresponding depth map (center), and ground truth labels (right).
 For visualization of the depth, we normalized and cropped the distance ranges per scan 
 and show the resulting values in false color.
 Best viewed in color on screen with zoom
 }
 \label{fig:Examples}
\end{figure*}

\replaced{An}{Four} example surface of the dataset is shown in Fig.~\ref{fig:Examples}. 
We depict the orthophoto, the corresponding depth map and the 
ground truth labels. Note that the peckings are sometimes virtually 
unrecognizable from the orthophoto and can hardly be discovered without taking
the ground truth labels into account. Further note the strong variation in depth 
ranges which stems from the shape and curvature of the rock surfaces themselves.

\section{Experiments}
\label{sec:Experiments}
In this section we present \added{baseline} experiments for our dataset. 
\deleted{that should serve as first baselines for surface segmentation}
We have published some complementary results on the 
dataset previously~\cite{Zeppelzauer2015dh} where we
focused on interactive segmentation and 
different types of hand-crafted surface features.
In contrast \replaced{to our previous work, in this paper}{, here} we focus on fully 
automatic segmentation and learned features. Aside 
from providing an evaluation protocol and\deleted{ first} baselines of state-of-the-art 
approaches we investigate the following questions related 
to our dataset in detail: (i) What is the benefit of using 
3D depth information compared to pure texture information (RGB) for 
surface segmentation of petroglyphs? (ii) Can our learned 
models generalize from rock surfaces of one location to 
surfaces of another location (generalization ability)? 


\subsection{Evaluation Protocol}
\label{sec:Protocol}
To enable reproducible and comparable experiments, we propose 
the following two evaluation protocols on the dataset:

\emph{4-fold Cross-Validaion:} To obtain results for the whole dataset, 
we perform a $k$-fold cross-validation,
with the number of folds being $k=4$. 
We randomly assigned the 
surface reconstructions to the folds. The assignment of surfaces to folds is provided with the dataset.

\emph{Cross-Site Generalization:} Here we separate the dataset into two sets 
according to the geographical locations the scans 
were captured at. We employ one of the two sets as 
training set and the other one as test set, 
and vice-versa.
In this way, we obtain insights about the generalization ability of 
a given approach across data from different capture locations.

The latter protocol is especially interesting 
since, on the one hand, the rock surfaces vary 
between sites, and on the other hand, the petroglyphs at different sites 
exhibit different shapes and peck styles, 
\eg, due to different tools that were used for their creation.
We separate the dataset into one set containing the scans 
from \emph{Seradina} and the other one containing 
the scans from \emph{Foppe di Nadro} and \emph{Naquane}.
Foppe di Nadro and Naquane were joined because these sites are situated 
next to each other 
and the corresponding petroglyphs are rather similar.
For evaluation we use one of the two sets as training set 
and the other one as test set, and vice-versa.
This results in the following three experiments:
\begin{itemize}
  \item Training on data from Foppe di Nadro and Naquane; testing on Seradina.
  \item Training on data from Seradina; testing on Foppe di Nadro.
  \item Training on data from Seradina; testing on Naquane.
\end{itemize}
In this way each surface reconstruction is exactly once in the 
test set. 

\subsubsection*{Metrics}
For quantitative evaluations on our dataset we propose a number of metrics commonly 
used for semantic segmentation to enable reproducible experiments\footnote{We 
provide the evaluation source code with the dataset}. 
In our case the segmentation task is a pixelwise binary problem
and, hence, the evaluation is based on 
the predicted segmentation mask and the ground truth mask.
Based on these masks we compute the Jaccard index~\cite{Jaccard1912}, 
also often termed region based intersection over union 
($IU$), for which we compute the average over classes ($mIU$) 
as in~\cite{Long2015cvpr,Hariharan2015cvpr,Zheng2015iccv,Kontschieder14pami}, 
the pixel accuracy (PA)~\cite{Shelhamer16pami,Kontschieder14pami}, 
the dice similarity coefficient ($DSC$)~\cite{zeppelzauer2016jooch},
the hit rate ($HR$)~\cite{Kontschieder14pami,zeppelzauer2016jooch} and the 
false acceptance rate ($FAR$)~\cite{zeppelzauer2016jooch}.

\subsection{Methods}
\label{sec:Methods}
To provide a baseline we evaluate the performance 
of prominent state-of-the-art approaches for semantic segmentation 
on our dataset. First, we perform 
experiments with a segmentation method based on Random Forests (RF). 
Second, we apply Convolutional Neural Networks (CNNs) 
\cite{LeCun1998ieee,Krizhevsky2012nips}, 
which currently show best performance on standard semantic segmentation benchmarks
\cite{Long2015cvpr,Hariharan2015cvpr,Zheng2015iccv,Chen2015iclr,Lin2016cvpr}
and compare them with the RF-based approach. We have shown previously that surface segmentation with 3D descriptors computed directly from the 3D point clouds is computationally demanding and with current state-of-the-art methods not performing well, see \cite{mzz15} for respective results for a subset of our dataset.  
Hence, we employ the depth maps \added{and orthophotos} generated from the point clouds as input to segmentation\deleted{ as well as the orthophotos}. 

For Random Forests (RFs) we employed an approach, 
which was also used as a baseline in many other RF-based works 
on semantic segmentation~\cite{Laptev2014gcpr,Kontschieder13cvpr,RotaBula2014cvpr}. 
That is, we trained a classification forest \cite{Breiman01} to compute a 
pixelwise labeling of the scans. 
The Random Forest is trained on patches representing the spatial neighborhood 
of the corresponding pixel.
To this end, we downscaled the scans by a factor of five
and extracted patches of size $17 \times 17$ 
corresponding to a side length of 6.8~mm.
We randomly sampled 8000 patches -- balanced over the classes -- from each training image.
As features we used the color or depth values directly. 
For all experiments we trained 10 trees, for which we stopped training when
a maximum depth of 18 was reached or less than a minimum number 
of 5 samples arrived in a node.

In the CNN-based approach we employ fully convolutional 
neural networks as proposed in \cite{Long2015cvpr}, 
since this work has been very influential for several 
following CNN-based methods for 
semantic segmentation \cite{Chen2015iclr,Zheng2015iccv,Lin2016cvpr}.
To perform petroglyph segmentation on our dataset
we finetune a model, which was pre-trained for semantic segmentation 
on the PASCAL-Context dataset \cite{Mottaghi2014cvpr}. 
To create training data for finetuning we again downscaled the depth maps 
by a factor of 5 and 
randomly sampled $224\times224$ pixel crops. 
To generate enough training data for finetuning the CNN and additionally 
increase the variation in the training set, we augment it with randomly  
rotated versions of the depth maps ($r \in \{ 0, 45, 90, \ldots, 315 \}$ degrees) 
prior to sampling patches.
Similarly, we flip the depth-maps with a probability of 0.5.
Note, that rotating the images randomly is reasonable since the  petroglyphs 
have no unique orientation on the rock surfaces. Using the described 
augmentation strategy we sampled about 5000 crops, while ensuring 
that each crop contains 
pixel labels from both classes.
We finetuned for a maximum of 30 epochs.
For finetuning we employ Caffe \cite{jia2014caffe} and set the 
learning rate to $5 \times 10^{-9}$.
Due to GPU memory limitations (3GB) we were only able to use a batch size of one 
(i.e. one depth map at a time).
We, thus, follow \cite{Shelhamer16pami} and use a high momentum of $0.98$, 
which approximates a higher batch size and might also yield better accuracy 
due to the more frequent weight updates \cite{Shelhamer16pami}.


\subsection{2D vs. 3D Segmentation}
\label{subsec:2dvs3d}
\begin{figure}[t]
\centering
\includegraphics[width=\columnwidth]{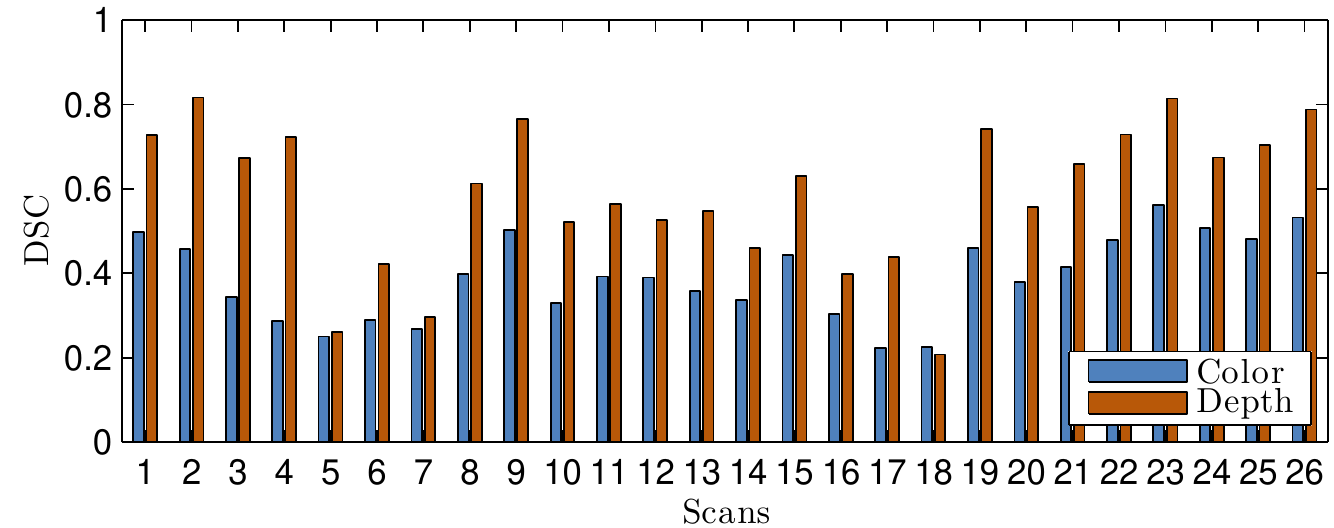}
\caption{
Dice Similarity Coefficient ($DSC$) per scan
}
\label{fig:ColorVsDepthPerImage}
\end{figure}

In a first experiment we investigate the importance 
of 3D information provided by our dataset compared to 
pure color-based surface segmentation.
Therefore, we train a Random Forest (RF) only with color information from the orthophotos 
and compare the results to a RF trained on only depth information.
For this experiment we follow the 4-fold cross-validation protocol specified 
in Sec.~\ref{sec:Protocol}.
The results in Tab.~\ref{tab:resultsColorVsDepth} (rows 1 and 2) clearly show the 
necessity for 3D information to obtain good results.
This is further underlined in Fig.~\ref{fig:ColorVsDepthPerImage}, where the results 
are compared for each individual scan. We observe that depth information improves 
results nearly for each scan by a large margin.
This can be explained by the fact that 
engraved surface regions often resemble 
the visual appearance of the surrounding 
rock surface due to influences 
from weathering.

\begin{table}[t]
  \centering
	\caption{
	Quantitative results for different setups, comparing the capabilities 
	of color (2D) and depth (3D) information. 
	3D segmentation strongly 
	outperforms color-based 2D segmentation
	}
	{
	  \begin{tabular}{ l c c c c c c }
	    \toprule
	    Representation	&  HR		& FAR 		& DSC		& mIU 		& PA \\
	    \midrule
	    Color		& 0.493 	& 0.675 	& 0.392 	& 0.465 	& 0.715 \\
	    Depth		& \textbf{0.779}& \textbf{0.553}& \textbf{0.568}& \textbf{0.569}& \textbf{0.779} \\
	    \midrule
	    Depth -- Cross-Sites	& 0.777 & 0.574 	& 0.550 	& 0.551 	& 0.763 \\
	    \bottomrule
	  \end{tabular}
  }
  \label{tab:resultsColorVsDepth}
\end{table}

Note that we also experimented with combining color and depth information, 
as well as with different features like image gradients,  
LBP features \cite{Ojala99pr}, 
and Haralick features \cite{Haralick1973} to abstract the pure color and depth information.
However, these had little to no impact 
on the final segmentation performance
and, hence, the results are omitted for brevity.


\subsection{Baseline Results}
In this section we present the results of the baseline methods 
for the two proposed evaluation protocols.

\subsubsection{Cross-Site Generalization}
The results for Random Forests 
for the proposed cross-site evaluation protocol (see Sec.~\ref{sec:Protocol})
are listed in Tab.~\ref{tab:resultsCrossvalidationOverRocks}.
Here, we provide the detailed results for each of the three splits.
Overall results 
averaged over all three experiments are shown in 
Tab.~\ref{tab:resultsColorVsDepth} (last row) for comparison with 
the experiments in Sec.~\ref{subsec:2dvs3d}. Interestingly, 
the overall results are in the same range as the results of 
the 4-fold cross-validation with randomly selected folds.
This suggests that -- using 3D information -- an automatic method 
is able to generalize from one site of the valley to another.

\begin{table}[t]
  \centering
	\caption{
	Results for cross-validation over different sites. 
	Quantitative results obtained for scans from \emph{Seradina} 
	when an RF classifier is trained on scans of only \emph{Foppe di Nadro} 
	and \emph{Naquane}, 
	as well as results for scans from 
	\emph{Foppe di Nadro} and \emph{Naquane} 
	when the classifier is trained only on scans of \emph{Seradina}
	}
	{
	  \begin{tabular}{ l c c c  }
	    \toprule				
	    Training Set:	& Foppe di Nadro $+$ Naquane 	& Seradina		& Seradina  \\
			\midrule
	    Test Set:		& Seradina 			& Foppe di Nadro	& Naquane \\
	    \midrule	    
	    HR 			& 0.843				& 0.706			& 0.744	 \\
	    FAR			& 0.544				& 0.274			& 0.644	 \\
	    DSC 		& 0.592				& 0.716			& 0.482	 \\
	    mIU 		& 0.612				& 0.704			& 0.446	 \\
	    PA 		  & 0.827				& 0.875			& 0.645	 \\
	    \bottomrule
	  \end{tabular}
	}
  \label{tab:resultsCrossvalidationOverRocks}
\end{table}

\subsubsection{4-fold Cross-Validation}
To provide a more comprehensive baseline for the performance 
of state-of-the-art methods we compare the results obtained 
with Random Forests (RFs) and 
Convolutional Neural Networks (CNNs) both evaluated on depth information.
For the CNN, which was pre-trained on color images (see Section~\ref{sec:Methods}) 
we simply fill all three input channels with the same depth 
channel to obtain a compatible input format.
Additionally, we subtract the local average depth 
value from each pixel in the depth map to normalize the input data,
which was necessary to stay compatible to the CNN pre-trained on RGB data. 
This normalization can be efficiently performed in a pre-processing step 
by subtracting a smoothed version of the depth map 
(Gaussian filter with $\sigma=\unit[12.5]{mm}$) from the depth map. 
This operation results in a local constrast equalization across 
the depth map \cite{mzz15} that better enhances the fine geometric 
details of the surface texture.

\begin{figure}[t]
\centering
\includegraphics[width=\columnwidth]{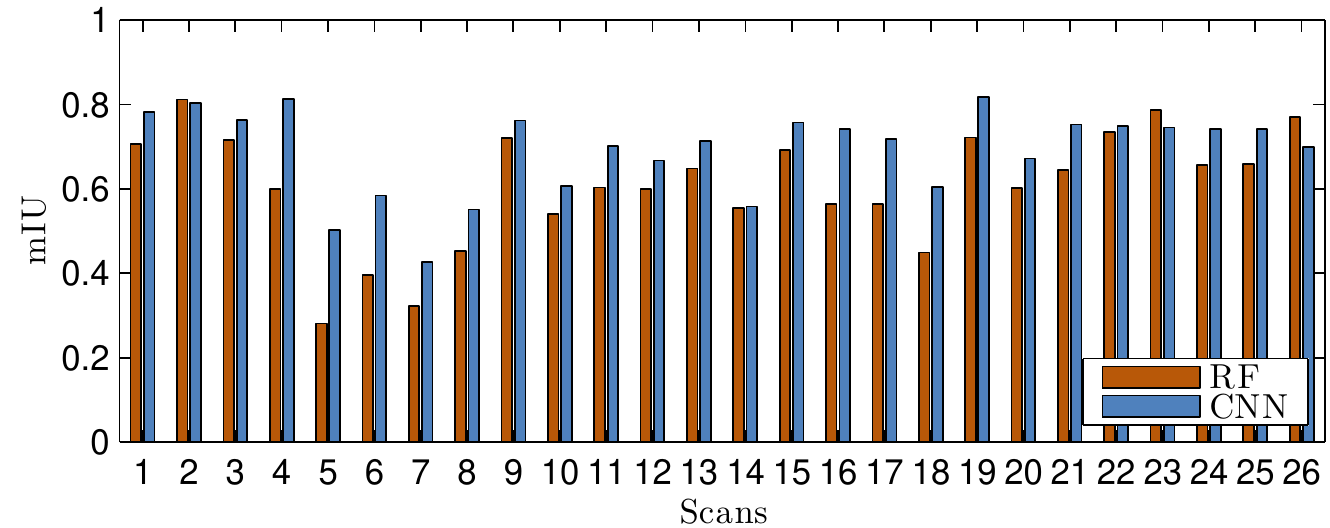}
\caption{
Mean intersection over union ($mIU$) per scan
}
\label{fig:BaselinesPerScan}
\end{figure}
\begin{figure*}[t]
\centering
 \subfloat[\scriptsize RGB]
  {\includegraphics[width=0.18\textwidth]{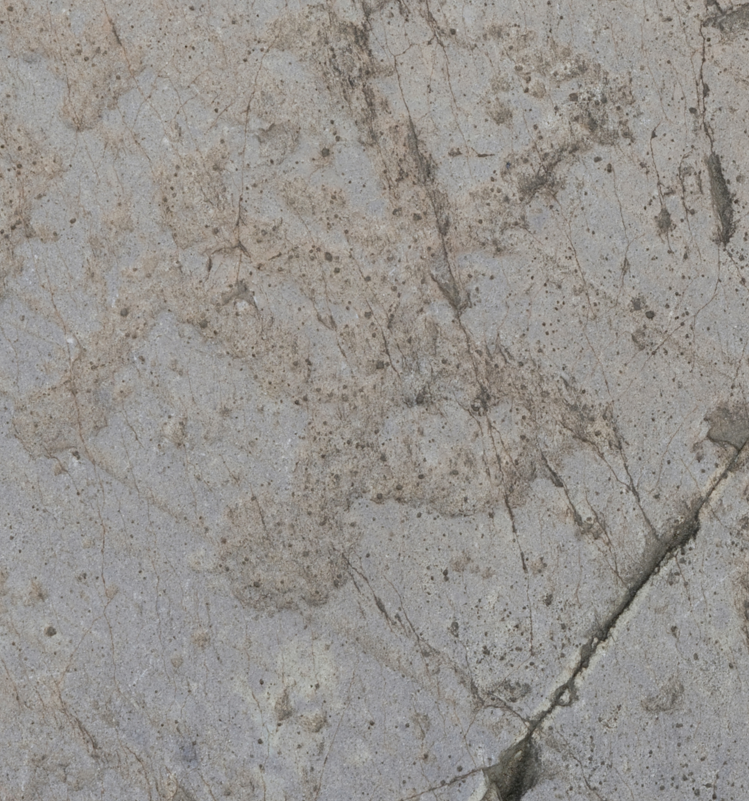}}
 \quad
 \subfloat[\scriptsize Depth map]
  {\includegraphics[width=0.18\textwidth]{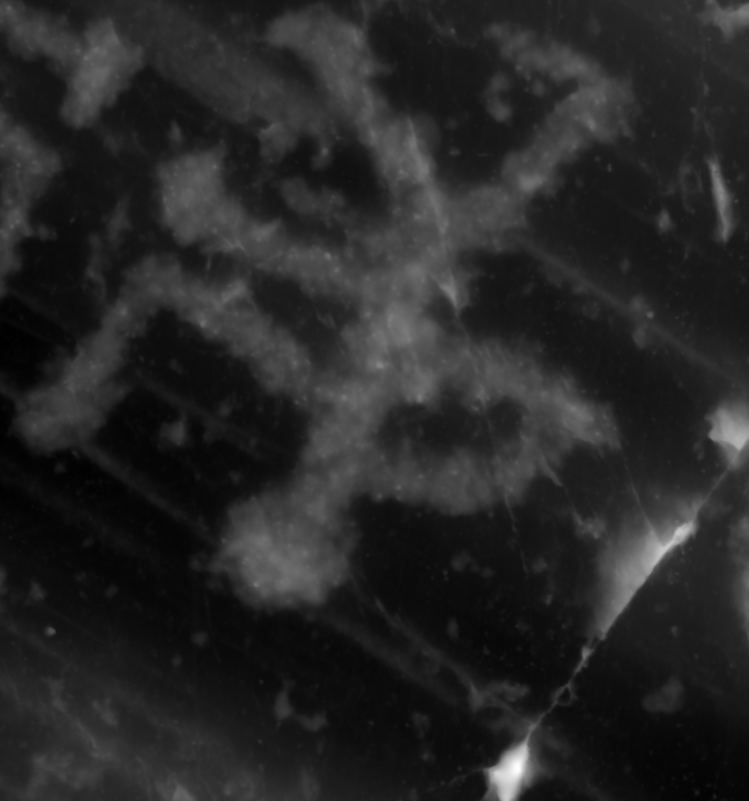}}
 \quad
 \subfloat[\scriptsize Ground truth]
  {\includegraphics[width=0.18\textwidth]{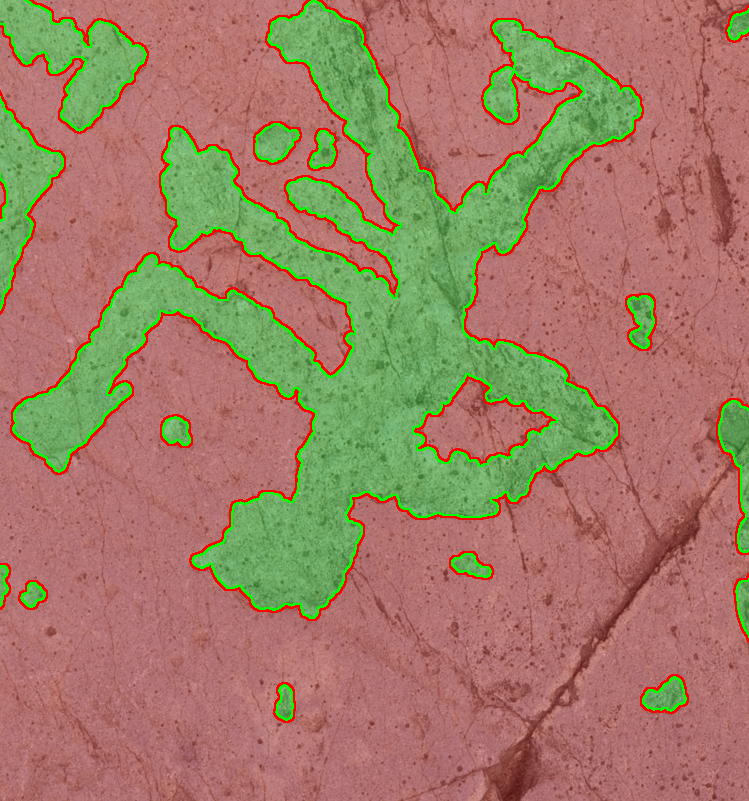}}
 \quad
 \subfloat[\scriptsize RF result]
  {\includegraphics[width=0.18\textwidth]{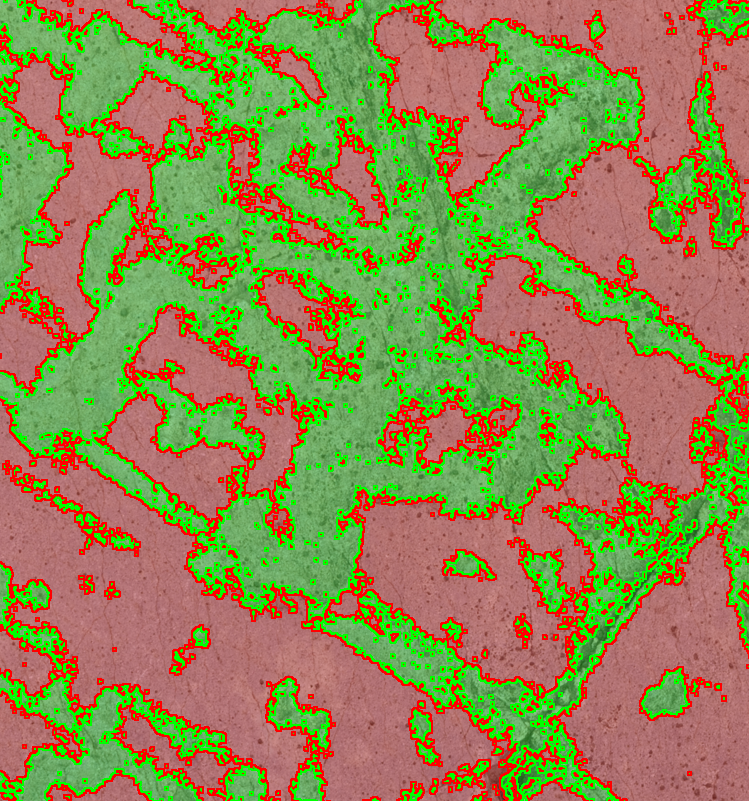}}
 \quad
 \subfloat[\scriptsize CNN result]
  {\includegraphics[width=0.18\textwidth]{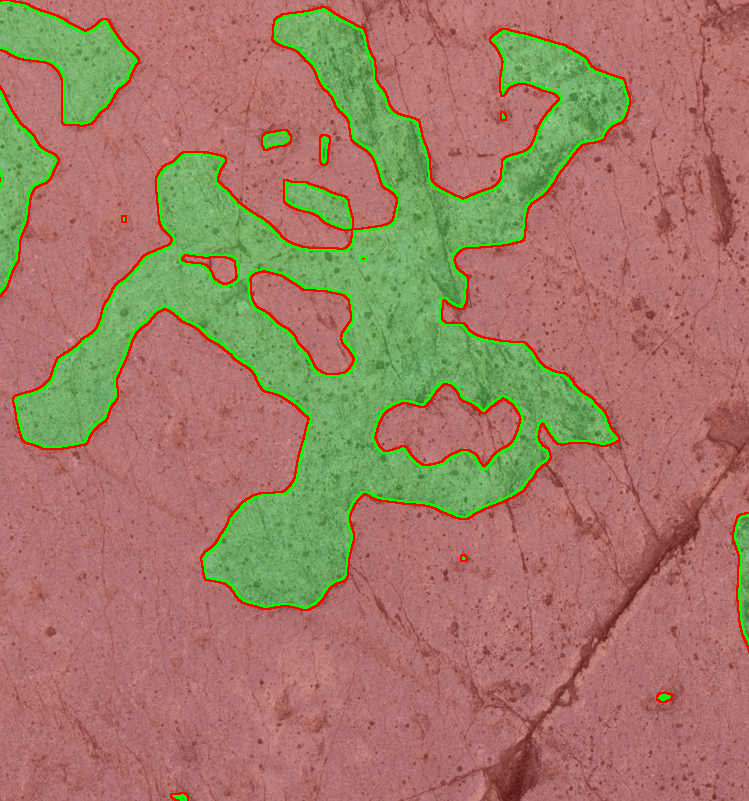}}
 \\
 \subfloat[\scriptsize RGB]
  {\includegraphics[width=0.18\textwidth]{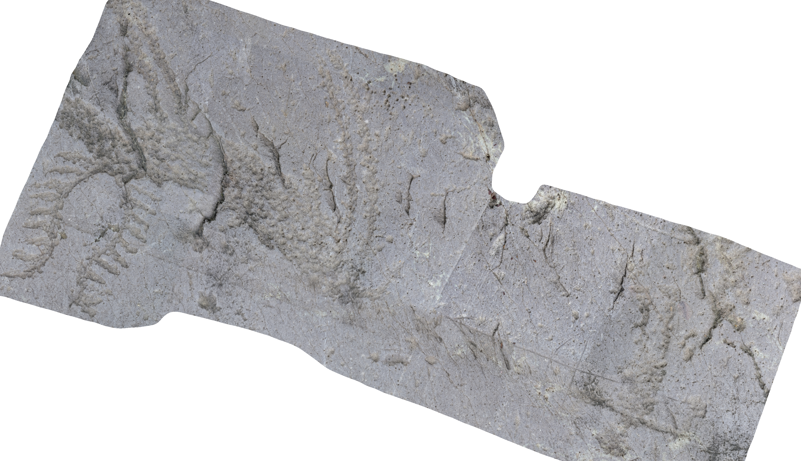}}
 \quad
 \subfloat[\scriptsize Depth map]
  {\includegraphics[width=0.18\textwidth]{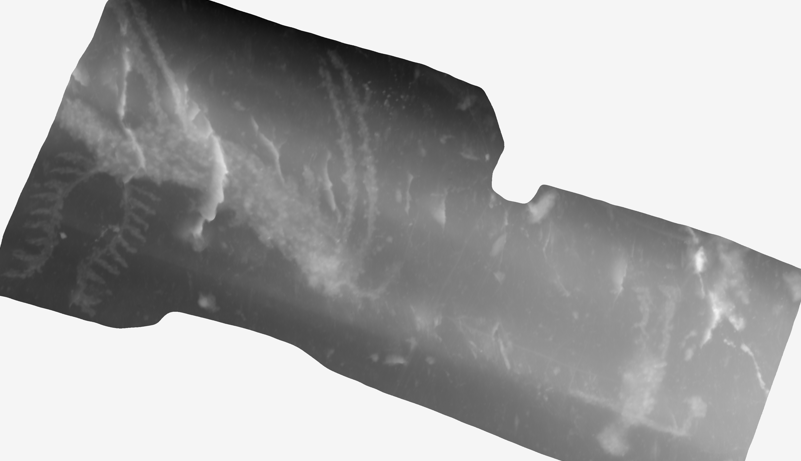}}
 \quad
 \subfloat[\scriptsize Ground truth]
  {\includegraphics[width=0.18\textwidth]{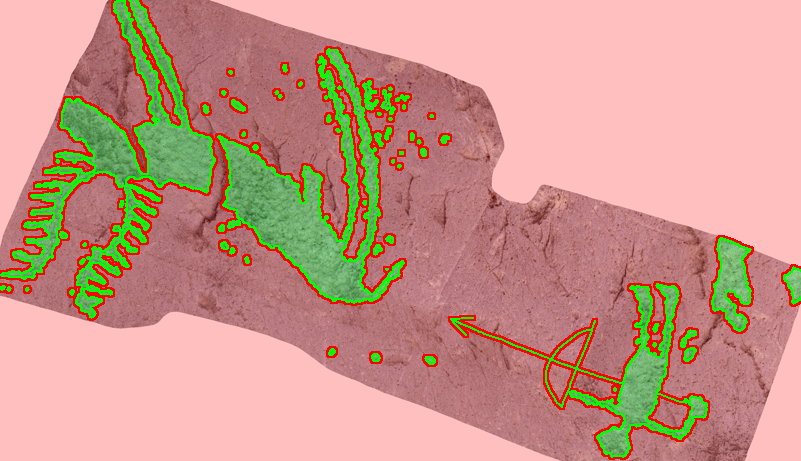}}
 \quad
 \subfloat[\scriptsize RF result]
  {\includegraphics[width=0.18\textwidth]{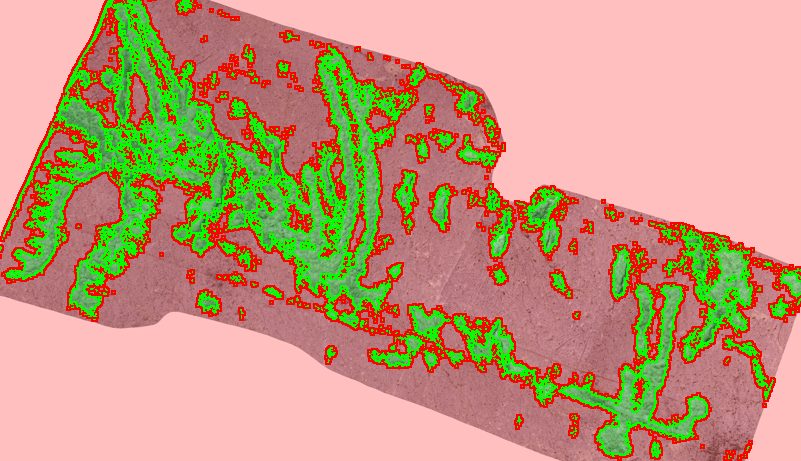}}
 \quad
 \subfloat[\scriptsize CNN result]
  {\includegraphics[width=0.18\textwidth]{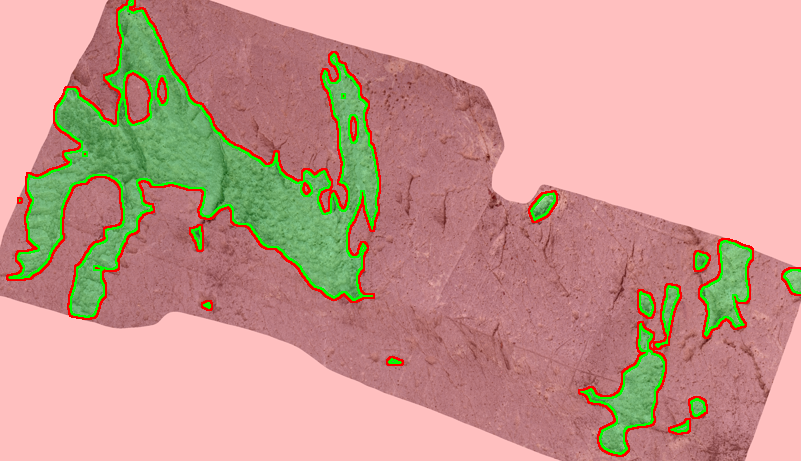}}
 \caption{
 Input images (orthophotos and depth maps), ground truth labelings and results 
 for the CNN and the RF baselines. 
 Best viewed on screen with zoom.
 }
 \label{fig:ExampleResults}
\end{figure*}

Quantitative results for the whole dataset are shown in 
Tab.~\ref{tab:ResultsBaselines}. 
The quantitative results  in terms of \emph{mIU} for each surface are 
visualized in Fig.~\ref{fig:BaselinesPerScan}.
In Fig.~\ref{fig:ExampleResults} we show some qualitative results 
for each method. 
From the results we observe that the Random Forest (RF) yields more cluttered results, 
whereas the the CNN yields more consistent but coarser segmentations.
The RF correctly detects small and thin pecked regions, which the CNN misses,
whereas the CNN usually captures the overall shape of the petroglyphs 
more accurately but misses details.
Note that for none of the results we applied Conditional or Markov Random Fields (MRFs, CRFs) or similar models, 
since we want to enable easier comparisons to our baselines.
We assume that the reasons for the differences of RF and CNN are (i) that the 
RF makes independent pixel-wise decisions whereas the CNN implicitly 
considers the spatial context through its learned feature hierarchy and (ii) 
that the receptive field of the RF is smaller than the receptive field 
of the CNN. This is because the CNN is able to exploit additional 
spatial information through its hierarchy of filters 
while the RF was unable to effectively exploit larger receptive fields 
in our experiments.

The complementary abilities of RF and CNN are further reflected in the 
quantitative results in Tab. \ref{tab:ResultsBaselines}. The more 
consistent and coarser segmentations of the CNN yield a better overall 
segmentation result which is reflected by the higher \emph{DSC}, \emph{mIU}, and \emph{PA} 
values. For the foreground class in particular the \emph{HR} of 
RF outperforms that of CNN which means that a higher percentage of 
foreground pixels is labeled correctly. The reason for this is that 
CNN often misses larger portions of the pecked regions. 

\begin{table}[t]
  \centering
	\caption{
	4-fold cross-validation results for Random Forests (RFs) and 
	Convolutional Neural Networks (CNNs)
	}
	{
	  \begin{tabular}{ l c c c c c c }
	    \toprule
	    Method		&  HR		& FAR 		& DSC		& mIU 		& PA \\
	    \midrule
	    RF			& \textbf{0.779}& 0.553		& 0.568		& 0.569		& 0.779 \\
	    CNN			& 0.693 	& \textbf{0.357}& \textbf{0.667}& \textbf{0.676}& \textbf{0.871} \\
	    \bottomrule
	  \end{tabular}
  }
  \label{tab:ResultsBaselines}
\end{table}

\section{Conclusions}
\label{sec:conclusions}
In this paper, we introduced a novel dataset for 3D surface 
segmentation. The dataset contains reconstructions of natural 
rock surfaces with complex-shaped engravings (petroglyphs). 
The main motivation for contributing the dataset to the community is to foster, 
in general, research on the automated semantic segmentation of 3D 
surfaces and, in particular, the segmentation of petroglyphs 
as a contribution to the conservation of our cultural heritage. 
We complement the dataset with accurate expert-annotated 
ground-truth, an evaluation protocol and provide baseline results for 
two state-of-the-art segmentation methods. 

Our experiments show that (i) depth information -- as provided by our dataset -- is imperative 
for the generalization ability of segmentation methods and pure 2D segmentation is insufficient 
for this dataset; (ii) in most cases, the use of CNN classification outperforms 
RFs in terms of quantitative measures and, 
qualitatively, the CNN yields rougher but more consistent segmentations
than RFs. The obtained results (baseline DSC of 0.667) show that 
the dataset is far from being solved and thus represents a challenge 
for 3D surface segmentation in future.



\section*{Acknowledgment}
The work leading to these results has been carried out in the project 3D-Pitoti,
which is funded from the European Community’s Seventh
Framework Programme (FP7/2007-2013) under grant agreement no 600545; 2013-2016.



\bibliographystyle{IEEEtran}
\bibliography{_abbrv_,biblio}
%
%
%

\end{document}